\documentclass[11pt]{article}
\usepackage{enumerate}
\usepackage{multirow}
\usepackage{dblfloatfix}
\usepackage{afterpage}
\usepackage{tikz}
\usetikzlibrary{arrows.meta, positioning}
\usepackage{url} 
\usepackage{graphicx,amsfonts,amsmath,amssymb,bm,natbib,booktabs,colortbl,xcolor,float,comment,makecell,dcolumn,multirow}
\usepackage[colorlinks,
            linkcolor=red,
            anchorcolor=green,
            citecolor=blue]{hyperref}
\usepackage{algorithm, algorithmicx, algpseudocode}
\floatname{algorithm}{Algorithm} 
\usepackage{enumitem}
\usepackage{pifont}
\newcommand{\cmark}{\textcolor{green!60!black}{\ding{52}}}
\newcommand{\xmark}{\textcolor{red}{\ding{56}}}
\usepackage{placeins}
\usepackage{listings}
\newtheorem{definition}{Definition}

\definecolor{blue}{RGB}{0, 144, 178}
\definecolor{red}{RGB}{255,18,0}

\definecolor{PolyRed}{RGB}{160,35,55}
\definecolor{SufeRed}{HTML}{811C21}

\definecolor{UIUCOrange}{HTML}{FF552E}
\definecolor{UIUCBlue}{HTML}{13294B}
\definecolor{yellow}{RGB}{240,228,66}
\definecolor{MSU}{RGB}{24, 69, 59}
\definecolor{Maroon}{HTML}{8B008B}

\usepackage[preprint]{acl}

\usepackage{times}
\usepackage{latexsym}

\usepackage[T1]{fontenc}

\usepackage[utf8]{inputenc}

\usepackage{microtype}

\usepackage{inconsolata}

\title{DOS: Dependency-Oriented Sampler for Masked Diffusion Language Models}

 \author{Xueyu Zhou, Yangrong Hu, \and Jian Huang\\
Department of Data Science and AI \\The Hong Kong Polytechnic University, Hong Kong, China\\
\textit{\small xueyu.zhou@connect.polyu.hk},\quad \textit{\small  yangrong.hu@connect.polyu.hk},\quad  \textit{\small j.huang@polyu.edu.hk} }

\begin{document}
\maketitle
\begin{abstract}
Masked diffusion language models (MDLMs) have recently emerged as a new paradigm in language modeling, offering flexible generation dynamics and enabling efficient parallel decoding.
However, existing decoding strategies for pre-trained MDLMs predominantly rely on token-level uncertainty criteria, while largely overlooking sequence-level information and inter-token dependencies.
To address this limitation, we propose \textbf{D}ependency-\textbf{O}riented \textbf{S}ampler (DOS), a training-free decoding strategy that leverages inter-token dependencies to inform token updates during generation. Specifically, DOS exploits attention matrices from transformer blocks to approximate inter-token dependencies, emphasizing information from unmasked tokens when updating masked positions. Empirical results demonstrate that DOS consistently achieves superior performance on both code generation and mathematical reasoning tasks. Moreover, DOS can be seamlessly integrated with existing parallel sampling methods, leading to improved generation efficiency without sacrificing generation quality.
\end{abstract}

\section{Introduction}
Large language models (LLMs) have achieved remarkable progress in recent years, demonstrating strong performance in tasks such as code generation and mathematical reasoning \citep{achiam2023gpt,grattafiori2024llama,guo2025deepseek,yang2025qwen3}.
Built upon transformer architectures \citep{vaswani2017attention}, most existing LLMs are trained via next-token prediction and generate text in an autoregressive (AR) manner \citep{radford2018improving,radford2019language,brown2020language}.
Despite its strong effectiveness, the AR paradigm inherently enforces a strict left-to-right generation order, which limits parallel decoding and constrains the flexibility of generation dynamics \citep{xia2024unlocking,qin2025backtrack,li2025refusion}.
These limitations have motivated the exploration of alternative generation paradigms beyond AR modeling.

Inspired by the success of diffusion models in continuous domains \citep{ho2020denoising,nichol2021improved,jing2022torsional,esser2024scaling}, masked diffusion language models (MDLMs) have recently emerged as a new paradigm for text generation \citep{austin2021structured,shi2024simplified,sahoo2024simple,nie2025large,ye2025dream}.
MDLMs define a forward noising process that progressively masks tokens in a text sequence, and learn a reverse denoising process to reconstruct the original sequence by predicting its masked tokens.
Compared to AR models, MDLMs enable flexible generation orders and allow tokens to be predicted in parallel at each denoising step. Some recent studies \citep{arriola2025block,yang2025diffusion,wang2025diffusion} further extend MDLMs to the block-wise diffusion model, where the sequence is partitioned into multiple blocks, striking a balance between auto-regressive models and diffusion language models.

To further strengthen generation in pre-trained MDLMs, various decoding strategies have been proposed to restructure the decoding order of masked tokens \citep{chang2022maskgit,koh2024plm,nie2025large,kim2025train} and improve the efficiency of parallel decoding \citep{wu2025fast,kim2025train,ben2025accelerated}.
For example, \citet{kim2025train} analyzes the influence of generation order in MDLMs and proposes a new inference strategy, leading to significant improvements in generation quality.
Although these methods have achieved remarkable results, several aspects of the decoding process remain underexplored.

A key limitation is that existing decoding strategies for MDLMs are primarily based on the output logits of MDLMs and leverage uncertainty-based criteria to guide token updates \citep{chang2022maskgit,koh2024plm,kim2025train,nie2025large}.
These criteria mainly focus on token-level uncertainty, while lacking an explicit mechanism to capture sequence-level information, particularly inter-token dependencies.
As a result, they may amplify the discrepancy between marginal token distributions and the true joint distribution over sequences.

\begin{figure}
    \centering
    \includegraphics[width=\linewidth]{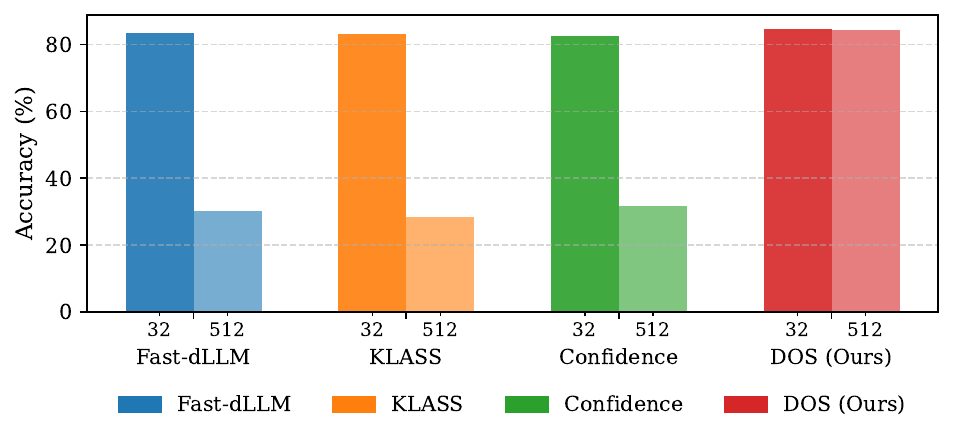}
    \caption{Accuracy on GSM8K using LLaDA-Instruct-8B \citep{nie2025large} with a fixed generation length of 512 tokens.
Block size 32 corresponds to block-wise decoding, while 512 represents the single-block setting. Existing methods (Fast-dLLM \citep{wu2025fast}, KLASS\citep{kim2025klass}, Confidence\citep{chang2022maskgit}) degrade under large block sizes, whereas DOS (ours) remains consistent and robust across both settings.}
    \label{fig1}
\end{figure}

In addition, some strategies introduce additional heuristic block structures that deviate from the original objective of MDLMs, possibly inducing a mismatch between training and inference \citep{wu2025fast,kim2025klass}.
When the imposed block partition is misaligned with the underlying dependency structure, it can disrupt the global organization of the sequence and make it difficult to generate samples consistent with the target joint distribution, leading to inconsistent and suboptimal generations.
As shown in Figure~\ref{fig1}, it makes generation sensitive to the choice of block size, potentially hindering its extension to the pure diffusion setting.
These observations highlight the need for decoding strategies that are independent of block structures and can explicitly model inter-token dependencies during generation.

From this perspective, we revisit the generation process of diffusion language models from a distributional perspective and analyze the role of decoding order in recovering the target joint distribution.
Based on this analysis, we propose \textbf{D}ependency-\textbf{O}riented \textbf{S}ampler (DOS), a training-free decoding strategy that leverages scores derived from the attention matrix to guide token updates. Specifically, we exploit the scaled dot-product attention weights, i.e., $softmax(QK^\top/\sqrt{d})$, as a proxy for inter-token dependency  \citep{vaswani2017attention}
, and use them to induce a dependency-aware decoding order.
This attention-based scoring mechanism enables us to recover dependencies between unmasked tokens and masked tokens, facilitating a dependency-aligned decoding order with respect to the token generation process.
Since the attention matrix can be directly obtained from the transformer block in the forward pass, our proposed method does not require any additional training or computational cost.
Empirically, DOS achieves high-quality generation in a single-block setting and still outperforms existing decoding strategies, even when those methods employ multiple decoding blocks.
Moreover, as a general criterion for decoding order, DOS can easily integrate other accelerated sampling methods, improving sampling efficiency while preserving generation quality.

\section{Related Works}
\paragraph{Discrete Diffusion Language Models}Early research into discrete diffusion generally falls into two paradigms: transition-based frameworks and score-based methods. D3PM~\citep{austin2021structured} pioneered the adaptation of continuous diffusion concepts~\citep{ho2020denoising} to discrete state spaces via corruption processes defined by transition matrices. Extending this to continuous time, \citet{campbell2022continuous} utilized continuous-time Markov chain (CTMC) theory to formulate the forward-backward dynamics, deriving a negative ELBO objective in the continuous limit. Alternatively, inspired by the success of denoising score matching~\citep{songscore}, several works have proposed discrete counterparts to the Stein score for modeling discrete data distributions~\citep{meng2022concrete, lou2023discrete}. More recently, the focus has shifted towards simplified MDLMs. Studies by \citet{ouyour, sahoo2024simple, shi2024simplified} demonstrate that simplified masking mechanisms can significantly enhance performance, effectively bridging the gap between diffusion-based and AR models. Transitioning from small-scale experiments to LLMs, recent works have begun to investigate the scaling laws of discrete diffusion. LLaDA~\citep{nie2025large} scales the architecture up to 8 billion parameters, showcasing impressive reasoning capabilities that were previously unseen in smaller discrete diffusion models like GPT-2~\citep{radford2019language}. Furthermore, Dream~\citep{ye2025dream} introduces a novel training paradigm by initializing diffusion models with pre-trained AR weights, thereby combining the strengths of both generative approaches.

\paragraph{Decoding Strategies} The decoding strategy that determines the order and pace of token generation is pivotal for the efficiency and quality of diffusion language models. While standard approaches like MDLM~\citep{sahoo2024simple} typically employ a stochastic unmasking schedule where masked tokens are updated randomly, this naive strategy is often suboptimal for complex reasoning tasks in large-scale models. To address this, recent research focuses on uncertainty-aware decoding. LLaDA~\citep{nie2025large} introduces a confidence-based criterion, prioritizing the unmasking of tokens with higher prediction confidence. Similarly, other works utilize entropy~\citep{koh2024plm} or margin confidence~\citep{kim2025train} as metrics to determine the decoding order. Beyond token-level ordering, LLaDA also explores a semi-autoregressive block-wise strategy, where sequences are generated in parallel blocks; this method has proven particularly effective for structured tasks like coding, significantly outperforming random ordering. Another critical line of research targets acceleration and dynamic scheduling. Instead of using a fixed top-$k$ selection, Fast-dLLM~\citep{wu2025fast} employs a flexible confidence threshold to adaptively select multiple tokens per step. To ensure generation stability, KLASS~\citep{kim2025klass} introduces a metric to measure the stabilization between consecutive predictions. Furthermore, the EB-sampler~\citep{ben2025accelerated} theoretically derives an upper bound for multi-token prediction errors, utilizing this bound to dynamically optimize the number of unmasked tokens at each step, thereby achieving a balance between speed and accuracy.

\begin{figure*}[!t]
    \centering

    \begin{minipage}[c]{0.3\textwidth}
        \centering
        \begin{tikzpicture}[
            scale=1.,
            transform shape,
            baseline=(current bounding box.center),
            node distance=1.2cm and 1cm,
            mynode/.style={
                circle,
                draw,
                minimum size=0.8cm,
                font=\small
            },
            >={Stealth[length=2mm]}
        ]
            \node[mynode] (C) {$C$};
            \node[mynode, below left=of C] (X1) {$X_1$};
            \node[mynode, below right=of C] (X2) {$X_2$};
            \node[mynode, below=of X1] (X3) {$X_3$};
            \node[mynode, below=of X2] (X4) {$X_4$};

            \draw[->, thick] (C) -- (X1);
            \draw[->, thick] (C) -- (X2);
            \draw[->, thick] (X1) -- (X3);
            \draw[->, thick] (X1) -- (X4);
            \draw[->, thick] (X2) -- (X4);
        \end{tikzpicture}
    \end{minipage}%
    \begin{minipage}[c]{0.5\textwidth}
        \footnotesize
        \setlength{\abovedisplayskip}{15pt}
        \setlength{\belowdisplayskip}{3pt}
        \begin{align*}
            \text{(a)}\quad & p(x_1, x_2, x_3, x_4 \mid C) \textcolor{red}{\neq}
            \underbrace{[p(x_1 \mid C)\, p(x_2 \mid C)\, p(x_3 \mid C)\, p(x_4 \mid C)]}_{\text{Step 1 \xmark}}
            \\ 
            \text{(b)}\quad & p(x_1, x_2, x_3, x_4 \mid C) =
            p(x_1, x_3 \mid C)\,
            p(x_2,x_4 \mid x_1, x_3, C)
            \\
            & \quad \textcolor{red}{\neq}
            \underbrace{[p(x_1 \mid C)\, p(x_3 \mid C)]}_{\text{Step 1 \xmark}}\,
            \underbrace{[p(x_2 \mid x_1, x_3, C)\,
            p(x_4 \mid x_1, x_3, C)]}_{\text{Step 2 \xmark}}
            \\ 
            \text{(c)}\quad & p(x_1, x_2, x_3, x_4 \mid C) =p(x_2, x_3 \mid C)\,
            p(x_1,x_4 \mid x_2, x_3, C)
            \\
            & \quad \textcolor{red}{\neq}
            \underbrace{[p(x_2 \mid C)\, p(x_3 \mid C)]}_{\text{Step 1 \cmark}}\,
            \underbrace{[p(x_1 \mid x_2, x_3, C)\,
            p(x_4 \mid x_2, x_3, C)]}_{\text{Step 2 \xmark}}
            \\ 
            \text{(d)}\quad & p(x_1, x_2, x_3, x_4 \mid C) =
            p(x_1, x_2 \mid C)\,
            p(x_3, x_4 \mid x_1, x_2, C)\\
           & \quad \textcolor{green!60!black}{=}\underbrace{[p(x_1 \mid C)\, p(x_2 \mid C)]}_{\text{Step 1 \cmark}}\,
            \underbrace{[p(x_3 \mid x_1, x_2, C)\,
            p(x_4 \mid x_1, x_2, C)]}_{\text{Step 2 \cmark}}\\
        \end{align*}
    \end{minipage}
    \caption{A toy example for parallel decoding in MDLMs, where $C$ denotes the prompt or unmasked tokens and $\{X_i\}_{i=1}^4$ are masked tokens.
    Different factorizations of $p(X_1,X_2,X_3,X_4|C)$ correspond to different parallel decoding orders.
    Only factorizations that respect the underlying dependency structure are able to recover the true joint distribution, while improper independence assumptions lead to distribution mismatch. }
    \label{fig2}
\end{figure*}

\section{Preliminaries}
In this section, we review the background and training objective of masked diffusion language models, and summarize their forward noising and reverse-time generation processes.

Let $\mathcal{V}=\{1,\ldots,V\}$ denote a discrete vocabulary of size $V$, and let $\bm{x}_0 \in \mathcal{V}^L$ denote a sequence of length $L$, where $\bm{x}_0^{i}$ is the $i$-th token.
To model the masking process, the state space is extended by introducing a dedicated mask symbol $\texttt{[M]} = V+1$, resulting in an augmented vocabulary of size $V+1$.
Let $\delta_a \in \mathbb{R}^{V+1}$ denote the one-hot vector whose $a$-th coordinate equals 1.

Masked diffusion language models can be formulated as a continuous-time masking (noising) process with factorized transitions
$q(\bm{x}_t|\bm{x}_s)=\prod_{i=1}^{L} q(\bm{x}_t^i|\bm{x}_s^i)$ for $0\le s<t\le 1$, where
$q(\bm{x}_t^i|\bm{x}_s^i)=\mathrm{Cat}\!\left(\bm{x}_t^i;\, \frac{\alpha_t}{\alpha_s}\delta_{\bm{x}_s^i}+\frac{\alpha_s-\alpha_t}{\alpha_s}\delta_{\texttt{[M]}}\right)$.
For example, LLaDA \citep{nie2025large} uses the linear schedule $\alpha_t = 1-t$, so $q(\bm{x}_t^i|\bm{x}_0^i)=\mathrm{Cat}\!\left(\bm{x}_t^i;\, (1-t)\delta_{\bm{x}_0^i}+t\delta_{\texttt{[M]}}\right)$ and the process reaches the fully masked state $\texttt{[M]}^L$ at $t=1$

Conditioned on $\bm{x}_0$, the posterior $q(\bm{x}_s \mid \bm{x}_t,\bm{x}_0)$ factorizes as $\prod_{i=1}^L q(\bm{x}_s^i \mid \bm{x}_t^i,\bm{x}_0^i)$, with
\begin{equation*}\label{eq:posterior-given-x0}
\resizebox{\linewidth}{!}{$
q(\bm{x}_s^i|\bm{x}_t^i,\bm{x}_0^i) = \left\{
\begin{array}{ll}
\mathrm{Cat}(\bm{x}_s^i; \delta_{\bm{x}_t^i}), & \bm{x}_t^i \neq \texttt{[M]},\\
\mathrm{Cat}\left(\bm{x}_s^i; \frac{1-\alpha_s}{1-\alpha_t}\delta_{\texttt{[M]}}+\frac{\alpha_s-\alpha_t}{1-\alpha_t}\delta_{\bm{x}_0^i}\right),& \bm{x}_t^i=\texttt{[M]}.\\
\end{array}
\right.
$}
\end{equation*}
Equation (\ref{eq:posterior-given-x0}) \citep{shi2024simplified, sahoo2024simple} suggests parameterizing the reverse-time transitions by substituting a learned predicted distribution of $\bm{x}_0$ into the analytic posterior:
\begin{align*} 
p_{\bm{\theta}}(\bm{x}_s|\bm{x}_t)&=q(\bm{x}_{s}|\bm{x}_t,\bm{x}_0=\bm{f}_{\bm{\theta}}(\bm{x}_t)) \nonumber \\
&=\prod_{i=1}^L q(\bm{x}_s^i|\bm{x}_t^i,\bm{x}_0^i=\bm{f}^i_{\bm{\theta}}(\bm{x}_t)),
\end{align*}
where $\bm{f}_{\bm{\theta}}(\bm{x}_t)\in \mathbb{R}^{L\times (V+1)}$ and $\bm{f}^i_{\bm{\theta}}(\bm{x}_t)\in \mathbb{R}^{V+1}$ is the model's output for the $i$-th token. $\bm{f}^i_{\bm{\theta}}(\bm{x}_t)$ satisfies $\sum_{j=1}^V\bm{f}^{i,j}_{\bm{\theta}}(\bm{x}_t)=1$ and $\bm{f}^{i,\texttt{[M]}}_{\bm{\theta}}(\bm{x}_t)=0$ for all $i$.
The training objective is to predict the masked states in each step with the $\bm{x}_0$, which can be simplified as :
\begin{equation*}\label{eq:loss}
    \resizebox{1.0\linewidth}{!}{$
    \mathcal{L}(\bm{\theta}) = -\mathbb{E}_{t,\bm{x}_0,\bm{x}_t}\left[\frac{1}{t}\sum_{i=1}^{L}\mathbf{1}[\bm{x}_t^{i}=\texttt{[M]}]\log p_{\bm{\theta}}(\bm{x}_0^{i}\mid \bm{x}_t)\right]
    $}.
\end{equation*}
By minimizing the training objective, $f_\theta$ can approximate the conditional distribution of masked tokens given $x_t$.
During inference, $\bm{x}_0^i$ is first predicted from $\bm{x}_t$ using $\bm{f}^i_{\bm{\theta}}(\bm{x}_t)$.
For masked states $\texttt{[M]}$, the predicted $\bm{x}_0^i$ is decoded with probability $\frac{\alpha_s-\alpha_t}{1-\alpha_t}$, while the token remains masked with probability $\frac{1-\alpha_s}{1-\alpha_t}$.
Unmasked states are kept unchanged.

\section{Distributional Analysis of Parallel Decoding}
\subsection{Distributional Mismatch in Parallel Decoding}

As discussed above, masked diffusion models are trained to match marginal distributions by conditioning on some unmasked tokens. However, the ultimate goal of the generation is to reconstruct the target joint distribution.

As illustrated by the toy example in Figure~\ref{fig2}, different parallel decoding strategies correspond to different factorizations of the joint distribution $p(X_1, X_2, X_3, X_4 \mid C)$.
Formulation~(a) predicts all tokens simultaneously and formulation~(b) performs parallel decoding under an incorrect block-wise factorization, which induces an invalid hierarchical dependency structure.
Simply imposing a block-wise partition does not resolve this issue, as the block structure itself may still be misaligned with the true dependency structure.

Under entropy-constrained decoding \citep{ben2025accelerated}, masked tokens can be selected at each step by explicitly exploring conditional independence among masked positions.
In formulation~(c), this strategy successfully identifies a subset of masked tokens, such as $X_2$ and $X_3$, that are conditionally independent given $C$. This enables an exact recovery of their joint distribution $p(X_2, X_3 \mid C)$ in the first step.
However, such conditionally independent subsets are not unique, and more importantly, the independence constraint is imposed only among masked tokens, while the dependency structure between the unmasked context $C$ and masked tokens is not explicitly considered.
As a result, although the joint distribution of the selected subset can be correctly recovered locally, the overall factorization remains inconsistent with the true dependency structure, ultimately preventing the recovery of the target joint distribution.

In contrast, formulation~(d) induces a factorization that is consistent with the underlying structure by explicitly accounting for the dependencies between the unmasked context and masked tokens.
In the first step, masked tokens $X_1$ and $X_2$, which are most strongly dependent on the available context $C$, are prioritized for parallel decoding.
Subsequently, the remaining tokens are decoded in parallel via conditioning on the newly unmasked ones.
Under such a dependency-consistent factorization, the target joint distribution can be correctly recovered through parallel decoding.

\subsection{Uncertainty-Based Decoding Methods}
In this subsection, we review and formalize a class of decoding strategies that rely on token-level uncertainty statistics computed from the model’s output distributions.
Denote the discrete distribution induced by the MDLM $f_\theta$ at time $t$ with input $x_t$ for position $i$ as $p_t^i=f^i_\theta(x_t)$.
\begin{definition}[Confidence ]\text{\citep{chang2022maskgit,nie2025large}}
For MDLM $f_\theta$ at time $t$ and position $i$, \textbf{Confidence} selects tokens based on the maximum value of the discrete distribution $p_t^i$ over the vocabulary $\mathcal{V}$, defined as
\[
\textit{conf}_t^i = \max_{v \in \mathcal{V}} p_t^i(v).
\]
\end{definition}

\begin{definition}[Entropy ]\text{\citep{koh2024plm}}
For MDLM $f_\theta$ at time $t$ and position $i$, \textbf{Entropy} measures the uncertainty of the discrete distribution $p_t^i$ over the vocabulary $\mathcal{V}$, defined as
\[
\textit{ent}_t^i = -\sum_{v \in \mathcal{V}} p_t^i(v)\log p_t^i(v).
\]
\end{definition}

\begin{definition}[Margin confidence]\text{\citep{kim2025train}}
For MDLM $f_\theta$ at time $t$ and position $i$, \textbf{Margin confidence} considers the difference between the highest and the second-highest probabilities in the discrete distribution $p_t^i$, defined as
\[
\textit{margin}_t^i = p_t^i(v^*) - \max_{v \in \mathcal{V}\setminus \{v^*\}} p_t^i(v),
\]
where $v^* = \arg\max_{v \in \mathcal{V}} p_t^i(v)$.
\end{definition}
Despite their differences, these decoding strategies all rely on token-level statistics extracted from the output distribution at individual positions.
Therefore, they can effectively identify confident or uncertain tokens locally and can recover marginal distributions for each masked token.
However, since methods do not explicitly account for dependencies across tokens, decoding strategies built upon them may fail to reconstruct the target joint distribution over the entire sequence.
This observation suggests that effective parallel decoding requires a principled strategy that goes beyond token-level uncertainty and selects tokens based on inter-token dependencies.

\begin{algorithm}[!t]
\caption{Attention-Based Dependency Scoring}
\label{alg1}
\begin{algorithmic}[1]
\Require Current sequence state $X_{t+1} \in \mathbb{R}^{L \times V}$, MDLM $f_\theta$, transformer block index $i$, masked position set $\mathcal{M}$
\Ensure Dependency score \textbf{\textit{dep}}$ \in \mathbb{R}^{L}$

\State $\text{outputs} \gets f_\theta(X_{t+1})$
\State $\mathrm{Attn}^{(i)} \gets \text{outputs.attentions}[i]$
\Comment{Multi-head attention weights from the $i$-th transformer block, $\mathrm{Attn}^{(i)} \in \mathbb{R}^{H \times L \times L}$}

\State $\mathrm{Attn} \gets \frac{1}{H}\sum_{h=1}^{H} \mathrm{Attn}^{(i)}_h$
\Comment{Head-averaged attention matrix $\mathrm{Attn} \in \mathbb{R}^{L \times L}$}

\State \textbf{\textit{dep}}$ \gets \mathbf{0} \in \mathbb{R}^{L}$
\State $\mathcal{U} \gets [L] \setminus \mathcal{M}$

\For{\textbf{each} $m \in \mathcal{M}$}
    \State $\textbf{\textit{dep}}(m) \gets \sum_{u \in \mathcal{U}} \mathrm{Attn}(m, u)$
    \Comment{Dependency of masked position $m$ on the unmasked context}
\EndFor

\State \Return Dependency score \textbf{\textit{dep}}
\end{algorithmic}
\end{algorithm}

\begin{table*}[!t]
\centering
\small
\resizebox{\textwidth}{!}{
\begin{tabular}{l l| c c c c c}
\toprule
Models & Methods &
\textbf{HumanEval} &
\textbf{MBPP } &
\textbf{GSM8K} &
\textbf{MATH500} \\
\midrule
\multirow{4}{*}{\makecell{LLaDA-Instruct-8B}} & Confidence
    & 27.44
    & 22.80
    & 31.69
    & 21.40  \\

& Entropy
    & 27.44
    & 19.40
    & 33.13
    & 18.60 \\
& Margin
    & 26.83
    & 26.40
    & 33.76
    & 22.20  \\
& DOS
    & \textbf{42.68 }
    & \textbf{38.40 }
    & \textbf{84.31}
    & \textbf{41.60}  \\
\midrule
\multirow{4}{*}{\makecell{Dream-v0-Instruct-7B}} & Confidence
    & 28.66
    & 39.60
    & 31.08
    & 13.60  \\
& Entropy
    & 25.61
    & 33.20
    & 30.55
    & 10.20\\
& Margin
    & 29.27
    & 42.80
    & 30.78
    & 15.20  \\
& DOS
    & \textbf{59.15}
    & \textbf{50.60}
    & \textbf{80.21}
    & \textbf{45.00} \\
\bottomrule
\end{tabular}}
\caption{Performance of different decoding strategies under top-1 sampling, in which all methods generate the entire sequence within a single block.
For the code benchmarks (HumanEval and MBPP), the generation length is fixed to 256 tokens, while for the math benchmarks (GSM8K and MATH500), the generation length is 512 tokens.}
\label{Tab1}
\end{table*}

\section{Methodology}

In this section, we propose \textbf{D}ependency \textbf{O}riented \textbf{S}ampler (DOS) for masked diffusion language models, which explicitly accounts for inter-token dependencies during generation.
DOS focuses on the problem of which masked tokens should be decoded at each denoising step.
Rather than determining the decoding order solely based on token-level uncertainty, our approach selects and updates masked tokens according to their dependency on the currently observed context.
In this work, we leverage the scaled dot-product attention weights, which explicitly model how information is aggregated across tokens
\citep{vaswani2017attention,clark2019does,voita2019analyzing,raganato2018analysis}, as an operational proxy for inter-token dependency.

Concretely, given the attention matrix extracted from a transformer block, each row is interpreted as describing how a query token integrates information from other tokens in the sequence.
For example, the entry at row $i$ and column $j$ of the attention matrix reflects how strongly the $i$-th token depends on information provided by the $j$-th token during the forward pass.
Since we focus on the information aggregated from unmasked tokens when predicting masked tokens, the dependency score for a masked token is defined as the total attention mass assigned to all unmasked tokens. Intuitively, a larger score indicates that the prediction at this masked position relies more heavily on the observed context, and should therefore be prioritized during decoding.

Formally, let $\mathcal{M}$ denote the set of masked positions and $\mathcal{U} = [L]\setminus\mathcal{M}$ the set of unmasked positions.
Given the multi-head attention weights $\mathrm{softmax}(QK^\top/\sqrt{d})$ extracted from a transformer block, we average across heads to obtain a token-to-token attention matrix $\mathrm{Attn} \in \mathbb{R}^{L \times L}$.
For each masked position $m \in \mathcal{M}$, the dependency score is defined as
\[
\textit{dep}(m) = \sum_{u \in \mathcal{U}} \mathrm{Attn}(m, u),
\]
which measures how strongly the prediction at position $m$ attends to the currently unmasked context.
In contrast to uncertainty-based criteria that are computed independently for each position,
the proposed score explicitly accounts for inter-token dependencies encoded in the attention structure.
The complete attention-based dependency scoring procedure is summarized in
Algorithm~\ref{alg1}.

Based on these scores, DOS selects a subset of masked tokens with the highest scores to decode at each step.
The decoding order is explicitly dependency-aware and better reflects
the dependency structure encoded by the model.
Since DOS provides a dependency score that ranks masked tokens, it can be directly combined with existing parallel decoding methods, such as top-K selection or entropy-constrained decoding (e.g., EB-Sampler).

\section{Experiments}

In this section, we present a comprehensive empirical evaluation of our proposed DOS across a diverse set of benchmarks and compare it against representative baseline decoding strategies.

\subsection{Datasets}
We evaluate DOS on a range of benchmarks covering different forms of structured generation, including code generation and mathematical reasoning:
\begin{itemize}
    \item \textbf{Code generation}: \textbf{HumanEval} \citep{chen2021evaluating} and \textbf{MBPP} \citep{austin2021program}, which consist of Python programming tasks with function-level specifications.

    \item \textbf{Mathematical reasoning}: \textbf{GSM8K} \citep{cobbe2021training}, which contains arithmetic word problems of grade school that require multi-step numerical reasoning and \textbf{MATH500} \citep{lightman2023let}, which is a subset of the MATH \citep{hendrycks2021measuring} benchmark and features more challenging problems.
\end{itemize}

\subsection{Baselines}
\begin{table*}[!htbp]
\centering
\small
\resizebox{\textwidth}{!}{
\begin{tabular}{l l| cc cc cc cc cc}
\toprule
&  &
\multicolumn{2}{c}{\textbf{HumanEval} } &
\multicolumn{2}{c}{\textbf{MBPP} } &
\multicolumn{2}{c}{\textbf{GSM8K}} &
\multicolumn{2}{c}{\textbf{MATH500}} \\
\cmidrule(lr){3-4}
\cmidrule(lr){5-6}
\cmidrule(lr){7-8}
\cmidrule(lr){9-10}
Models & Methods & Acc $\uparrow$ & NFE $\downarrow$
  & Acc $\uparrow$ & NFE $\downarrow$
  & Acc $\uparrow$ & NFE $\downarrow$
  & Acc $\uparrow$ & NFE $\downarrow$  \\

\midrule
\multirow{6}{*}{\makecell{LLaDA-Instruct-8B}}
& Fast-dLLM
    & 28.66 & 83.26
    & 23.00 & 78.93
    & 33.06 & 195.46
    & 20.80 & 219.54 \\
& KLASS
    & 29.88 & 107.99
    & 24.00 & 97.09
    & 28.43 & 265.52
    & 19.60 & 300.48 \\
& Confidence$+$EB
    & 28.66 & 160.10
    & 22.80 & 167.43
    & 32.15 & 299.59
    & 21.00 & 327.07 \\
& Entropy$+$EB
    & 29.88 & 175.71
    & 18.80 & 183.72
    & 25.09 & 309.12
    & 19.00 & 334.24 \\
& Margin$+$EB
    & 26.22 & 159.58
    & 26.60 & 166.00
    & 38.51 & 293.45
    & 21.40 & 324.10 \\
& DOS$+$EB
    & \textbf{45.12} & 115.84
    & \textbf{38.80} & 76.098
    & \textbf{84.61} & 163.48
    & \textbf{41.00} & 244.24 \\
\midrule
\multirow{6}{*}{\makecell{Dream-v0-Instruct-7B}}
& Fast-dLLM
    & 28.66 & 141.62
    & 39.60 & 127.69
    & 31.08 & 205.52
    & 13.60 & 273.58 \\
& KLASS
    & 27.44 & 150.46
    & 38.00 & 145.09
    & 31.84 & 239.92
    & 14.20 & 302.69 \\
& Confidence$+$EB
    & 28.66 & 199.66
    & 39.60 & 191.74
    & 31.01 & 302.89
    & 13.60 & 384.77 \\
& Entropy$+$EB
    & 25.61 & 222.15
    & 32.60 & 214.86
    & 30.63 & 356.59
    & 10.00 & 408.54\\
& Margin$+$EB
    & 29.27 & 183.25
    & 41.20 & 177.73
    & 30.86 & 297.96
    & 15.40 & 377.15 \\
& DOS$+$EB
    & \textbf{58.54} & 72.47
    & \textbf{51.00} & 68.33
    & \textbf{79.98} & 182.43
    & \textbf{45.40} & 229.36\\
\bottomrule
\end{tabular}}
\caption{Performance of different decoding strategies within a single block. \textbf{EB} stands for EB-sampler \citep{ben2025accelerated}, a parallel decoding method that can be combined with different scores. \textbf{Acc} denotes task accuracy, and \textbf{NFE} denotes the number of model forward evaluations, reflecting decoding efficiency. For HumanEval and MBPP, the generation length is fixed to 256 tokens, while for GSM8K and MATH500, the generation length is 512 tokens.}
\label{Tab2}
\end{table*}
We compare DOS against the following baselines:
\begin{itemize}
    \item \textbf{Uncertainty-based methods}: selecting tokens at each step based on uncertainty criteria, including confidence \citep{chang2022maskgit,nie2025large}, entropy \citep{koh2024plm}, and margin confidence \citep{kim2025train}.
    \item \textbf{Entropy-Based (EB) Sampler} \citep{ben2025accelerated}: selecting multiple tokens at each step under an entropy-constrained sampling scheme. EB-sampler can be easily integrated with other methods, such as confidence, entropy, margin confidence and our proposed method, by ranking tokens according to these criteria.
    \item \textbf{Fast-dLLM} \citep{wu2025fast}: selecting multiple tokens with confidence greater than a threshold at each step.
    \item \textbf{KLASS} \citep{kim2025klass}: selecting multiple tokens with thresholds on both confidence and token-level KL divergence, where the KL divergence measures the difference between the previous distribution and the current distribution of the given token.
\end{itemize}
We apply these methods to two open source MDLMs, including LLaDA-Instruct-8B \citep{nie2025large} and Dream-v0-Instruct-7B \citep{ye2025dream}.
Both models are trained with the masked diffusion objective, without utilizing block-wise masking structures.
Further implementation details are provided in Appendix~\ref{appendix:baselines} and hyperparameters are provided in Appendix~\ref{appendix:hyperparams}.

\subsection{Main Result}

\paragraph{\textbf{DOS significantly improves performance under single-block decoding.}}

We first evaluate DOS under a single-block decoding setting, where the entire sequence is generated within one diffusion block.
As shown in Table~\ref{Tab1}, DOS consistently outperforms uncertainty-based decoding strategies across all benchmarks and both MDLMs.
In contrast, confidence, entropy, and margin struggle to generate long sequences within a single block and exhibit degraded performance, particularly on mathematical reasoning tasks that require strict logical progression across extended generations.
By explicitly leveraging inter-token dependencies, DOS can capture long-range dependencies that govern reasoning structure and achieve substantial performance gains, improving accuracy by large margins on GSM8K and MATH500.
These results demonstrate that dependency-aware token selection is crucial for generating sequences consistent with the target joint distribution under single-block decoding.

\paragraph{\textbf{DOS integrates seamlessly with parallel decoding and improves efficiency without sacrificing accuracy.}}
We next evaluate whether DOS can be combined with parallel decoding strategies to improve efficiency.
Specifically, we integrate DOS with the EB sampler, which coordinates parallel updates among masked tokens under entropy constraints.
As shown in Table~\ref{Tab2}, DOS$+$EB consistently improves both accuracy and decoding efficiency for LLaDA and Dream models.
Compared with the results of EB sampler integrated with uncertainty-based methods, DOS not only improves the accuracy across tasks but also achieves at least a 1.3$\times$ speedup on the LLaDA model.
Similar trends are observed for the Dream model.
In addition, compared to alternative efficient decoding baselines such as Fast-dLLM and KLASS, DOS$+$EB achieves substantially higher accuracy while maintaining competitive decoding efficiency.
These results indicate that DOS provides a complementary, dependency-aware signal that enhances existing parallel decoding methods.

\begin{table}[!htbp]
\centering
\small
\resizebox{0.5\textwidth}{!}{
\begin{tabular}{l|cccc}
\toprule
\textbf{Method} &
\textbf{HumanEval} &
\textbf{MBPP} &
\textbf{GSM8K} &
\textbf{MATH500} \\
\midrule
Confidence
    & 40.85 & 38.20 & 82.56 & 38.80 \\
Entropy
    & 41.46 & 37.80 & 82.79 & 40.00 \\
Margin
    & 40.85 & 37.40 & 82.64 & 38.80 \\
Fast-dLLM
    & 41.46 & \underline{38.80} & \underline{83.40} & 39.60 \\
KLASS
    & \underline{42.07} & 38.60 & 83.02 & \underline{40.20} \\
\midrule
\textbf{DOS (w/o block) }
    & 42.68 (\textcolor{green!60!black}{+0.61})
    & 38.40 (\textcolor{red!70!black}{-0.40})
    & 84.31 (\textcolor{green!60!black}{+0.91})
    & 41.60 (\textcolor{green!60!black}{+1.40}) \\
\textbf{DOS (w/ block) }
    & \textbf{44.51} (\textcolor{green!60!black}{+2.44})
    & \textbf{38.80} (\textcolor{green!60!black}{+0.00})
    & \textbf{84.61} (\textcolor{green!60!black}{+1.21})
    & \textbf{42.40} (\textcolor{green!60!black}{+2.20}) \\
\bottomrule
\end{tabular}}
\caption{
Performance comparison on HumanEval, MBPP, GSM8K, and MATH500 using LLaDA-Instruct-8B.
All baseline methods (Confidence, Entropy, Margin, Fast-dLLM, and KLASS) adopt the block diffusion with a fixed block size of 32. DOS (ours) is evaluated both without block partitioning (single-block) and with block partitioning (block size = 32).
}
\label{tab3}
\end{table}

\paragraph{\textbf{DOS consistently outperforms block-based decoding strategies with and without block partitioning.}}

We further evaluate DOS against decoding strategies with block structures on the LLaDA model.
All baseline methods employ block diffusion with a fixed block size of 32, whereas DOS is evaluated both with and without block partitioning.
As shown in Table~\ref{tab3}, DOS consistently achieves superior performance across all benchmarks, even when block partitioning is removed.
Although block partitioning improves the performance of existing uncertainty-based decoding strategies, DOS consistently outperforms these methods under both block-based and single-block settings.
Moreover, DOS maintains strong performance without relying on block structures and further benefits when block diffusion is introduced.
These results demonstrate that explicitly modeling inter-token dependencies is more critical than the choice of block structure for achieving high-quality parallel decoding.
 We further conduct additional comparisons under block-based decoding with different configurations and details are presented in Appendix~\ref{B}.

\begin{figure}[!t]
    \centering
    \includegraphics[width=\linewidth]{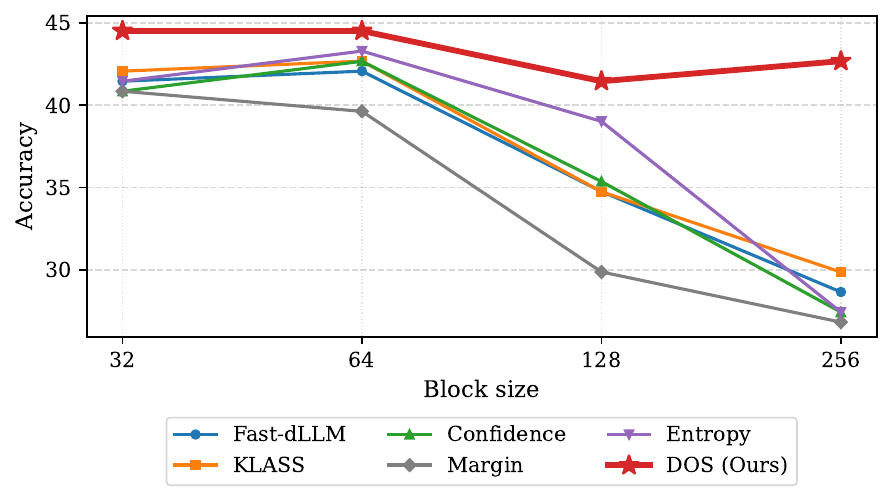}
    \caption{Accuracy on HumanEval using LLaDA-Instruct-8B with a fixed generation length of 256 under varying block sizes. Existing decoding strategies are sensitive to block size and degrade as the block size increases, whereas DOS (ours) demonstrates strong robustness to block size variation and maintains superior consistency across all settings.}
    \label{fig3}
\end{figure}

\paragraph{\textbf{DOS is robust to block size and consistently outperforms uncertainty-based decoding.}}

We further investigate the sensitivity of different decoding strategies to the choice of block size and the accuracy of various methods under different block sizes is reported in Figure~\ref{fig3}.
Existing methods are sensitive to the block configuration and exhibit noticeable performance fluctuations as the block size increases.
In contrast, DOS maintains consistently strong performance across a wide range of block sizes and consistently outperforms competing methods.
These results indicate that DOS provides a more stable and robust decoding signal by explicitly modeling inter-token dependencies, making it less sensitive to the block configuration. Additional results on the Dream model are provided in Appendix~\ref{B.2}

\begin{figure}[!htbp]
    \centering
    \includegraphics[width=\linewidth]{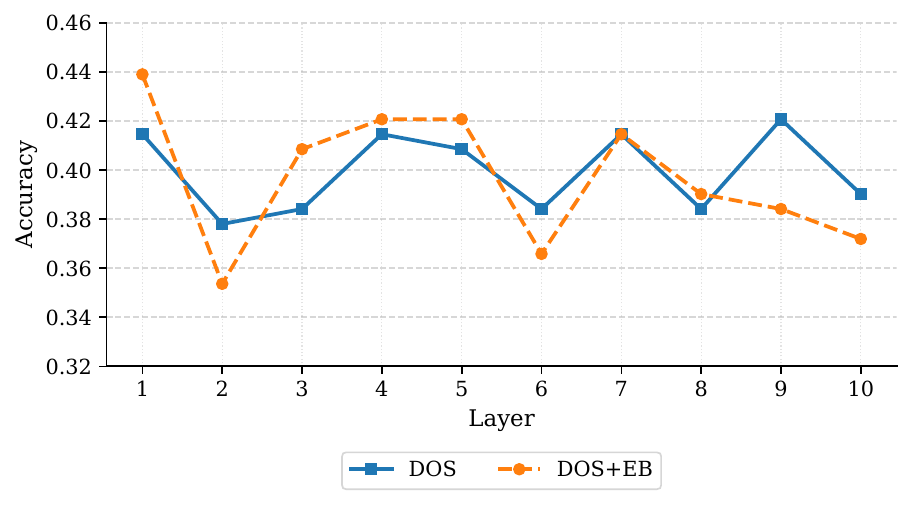}
    \caption{Accuracy of DOS and DOS+EB on the HumanEval benchmark using the LLaDA-Instruct-8B model, where the x-axis indicates the transformer layer from which the attention matrix is extracted.}
    \label{fig4}
\end{figure}

 \paragraph{\textbf{Effect of Transformer Layer Choice.}}

We further explore the role of layer depth in attention-based dependency scoring within the single-block setting. As illustrated in Figure~\ref{fig4}, the performance of DOS and DOS$+$EB on HumanEval remains robust across the first 10 layers. This phenomenon suggests that the shallow layers can capture the syntactic structures required for effective masked-token selection, which is consistent with prior studies \citep{tenney2019bert,clark2019does,jawahar2019does}.
Additional experiments are provided in Appendix~\ref{B.3}.

\section{Conclusion}

In this work, we study the decoding process of masked diffusion language models from a dependency-aware perspective and show that existing decoding strategies are limited by their reliance on token-level uncertainty and their sensitivity to heuristic block structures, both of which stem from a lack of explicit modeling of inter-token dependencies.
To address these limitations, we propose the Dependency-Oriented Sampler (DOS), a training-free decoding strategy that leverages attention-derived dependency signals to guide the decoding order of masked tokens.

By using the attention matrix as a proxy for inter-token dependencies, DOS updates tokens that are better supported by the current unmasked context, leading to a decoding process that is more aligned with the underlying joint distribution.
Empirical results on code generation and mathematical reasoning benchmarks demonstrate that DOS consistently outperforms existing uncertainty-based decoding strategies.
Notably, DOS can be seamlessly integrated with parallel sampling methods, improving generation efficiency without sacrificing quality.

Our analysis highlights the importance of respecting inter-token dependencies during sampling and suggests that the choice of decoding order plays a critical role in recovering the target joint distribution in masked diffusion language models.
We hope this work encourages further investigation into dependency-aware decoding strategies and contributes to a deeper understanding of parallel generation dynamics in diffusion language models.

\clearpage
\newpage
\section*{Limitations}
This work has several limitations that suggest directions for future research.

First, DOS relies on attention weights extracted from transformer blocks as a proxy for inter-token dependencies.
While attention provides an effective and readily available signal for guiding decoding order, it does not necessarily correspond to explicit causal or structural dependencies among tokens.
Some studies have explored incorporating explicit dependency structures, such as directed acyclic graphs (DAGs), into language model training to better capture compositional and dependency-aware generation dynamics \citep{huang2022DATransformer,huang2022PDAT}.
Integrating such structured dependency representations into the decoding process, or combining them with attention-derived signals, may provide a more principled alternative and is an interesting direction for future work.

Second, our current implementation extracts dependency scores from a single transformer layer with head-averaged attention.
Although we observe strong performance across different settings, this design does not fully exploit the hierarchical nature of representations in deep transformers.
Extending DOS to integrate information across multiple layers, or to selectively aggregate signals from different attention heads, could enable richer dependency modeling and further improve decoding performance.

Finally, our evaluation focuses on structured generation tasks, including code generation and mathematical reasoning, where long-range dependencies play a critical role.
The effectiveness of DOS in other generation settings, such as open-ended dialogue or multilingual generation, remains to be systematically investigated.
Exploring how dependency-oriented decoding interacts with different task characteristics is an important direction for future work.

\section*{Ethics Statement}

This work focuses on inference-time decoding strategies for pre-trained masked diffusion language models.
It does not involve human subjects, data collection, or additional model training.
All experiments are conducted on publicly available benchmarks.
The proposed method does not introduce new ethical risks beyond those inherent to the underlying language models.

\bibliography{dos1}

\clearpage
\appendix
\newpage
\noindent
\textbf{\Large Appendix}

\section{Experiment Details}
\label{sec:appendix_experiment}

\subsection{Experimental Setup}
\label{appendix:exp_setup}
We conduct all experiments with LLaDA-Instruct-8B \citep{nie2025large} and Dream-v0-Instruct-7B \citep{ye2025dream}. We run inference with batch size of 1 on a single \textbf{NVIDIA L40} GPU with 48 GB memory. For deterministic decoding, we set temperature to 0 and random seed to \textbf{42} for all runs.
\subsection{Baseline Methods}
\label{appendix:baselines}
This subsection provides implementation details of the key baseline methods used in our experiments, including uncertainty-based methods and accelerated samplers. Denote the discrete distribution induced by the MDLM $f_\theta$ at time $t$ with input $x_t$ for position $i$ as $p_t^i=f^i_\theta(x_t)$.

\paragraph{Confidence}\text{\citep{chang2022maskgit,nie2025large}}
For MDLM $f_\theta$ at time $t$ and position $i$, \textbf{Confidence} selects tokens based on the maximum value of the discrete distribution $p_t^i$ over the vocabulary $\mathcal{V}$, defined as
\[
\textit{conf}_t^i = \max_{v \in \mathcal{V}} p_t^i(v).
\]
\paragraph{Entropy}~\citep{koh2024plm} For MDLM $f_\theta$ at time $t$ and position $i$, \textbf{Entropy} measures the uncertainty of the discrete distribution $p_t^i$ over the vocabulary $\mathcal{V}$, defined as
\[
\textit{ent}_t^i = -\sum_{v \in \mathcal{V}} p_t^i(v)\log p_t^i(v).
\]
\paragraph{Margin confidence}~\citep{kim2025train} For MDLM $f_\theta$ at time $t$ and position $i$, \textbf{Margin confidence} considers the difference between the highest and the second-highest probabilities in the discrete distribution $p_t^i$,, defined as
\[
\textit{margin}_t^i = p_t^i(v^*) - \max_{v \in \mathcal{V}\setminus \{v^*\}} p_t^i(v),
\]
where $v^* = \arg\max_{v \in \mathcal{V}} p_t^i(v)$.

\paragraph{Fast-dLLM}
Fast-dLLM performs confidence-based parallel decoding by unmasking all positions whose confidence exceeds a confidence threshold $\epsilon$.
Concretely, at each step $t$, let $\mathcal{M}_t$ denote the set of masked positions. It computes the confidence score $\text{conf}_t^i$ and masks the subset of positions
\begin{equation*}
    \begin{aligned}
        \mathcal{S}_t &= \left\{ i \in \mathcal{M}_t \,\middle|\, \text{conf}_t^i = \max_{v \in \mathcal{V}} p_t^i(v) > \epsilon \right\}.
    \end{aligned}
\end{equation*}
Compared to fixed-budget top-$K$ unmasking, this approach yields a dynamic number of updates per iteration and can be integrated with semi-autoregressive block decoding for enhanced efficiency. Furthermore, while Fast-dLLM supports block-wise KV cache to eliminate redundant computations of key and value vectors across previous blocks, our evaluation focuses on decoding strategy rather than hardware-specific wall-clock speed; consequently, we do not utilize block-wise KV cache in our experiments.
\paragraph{KLASS.}
KLASS performs stable-token selection by identifying positions where the predictive distribution is stable across steps and reaches high confidence.
At each step $t$, let $\mathcal{M}_t$ denote the set of masked positions. KLASS selects a subset of stable tokens $\mathcal{S}_t \subseteq \mathcal{M}_t$ to be unmasked based on a history length $n$, a KL-divergence threshold $\epsilon_{\text{KL}}$, and a confidence threshold $\tau$:
\begin{equation*}
    \begin{aligned}
        \mathcal{S}_t = & \Big\{ i \in \mathcal{M}_t \;\Big|\;
        \forall k \in \{1, \dots, n\}, \\
        &D_{\text{KL}}(p_{t-k}^i \,\|\, p_{t-k+1}^i) < \epsilon_{\text{KL}}
        \wedge\; \text{conf}_t^i > \tau \Big\},
    \end{aligned}
\end{equation*}
where $D_{\text{KL}}$ denotes the Kullback-Leibler divergence between the predictive distributions of consecutive steps at position $i$, and $\text{conf}_t^i = \max_{v \in \mathcal{V}} p_t^i(v)$ is the confidence score.
By requiring predictions to remain consistent over $n$ steps, KLASS ensures that only stable tokens are updated in parallel, thereby improving generation reliability.

\paragraph{EB-Sampler.}
The entropy-bounded sampler accelerates generation by unmasking a number of tokens per step that satisfy the entropy constraint.
Concretely, at each step $t$, it first ranks all masked positions in $\mathcal{M}_t$ based on a specific score like confidence \textit{conf} or our proposed dependency  \textit{dep}.
Then, starting from the highest-ranked token, it selects the largest subset $\mathcal{S}_t \subseteq \mathcal{M}_t$ that satisfies the cumulative entropy constraint:
\begin{equation*}
    \sum_{i \in \mathcal{S}_t} \text{ent}_t^i  - \max_{j \in \mathcal{S}_t} \text{ent}_t^j \le \gamma,
\end{equation*}
where $\text{ent}_t^i = -\sum_{v \in \mathcal{V}} p_t^i(v)\log p_t^i(v)$ denotes the entropy of the predictive distribution at position $i$.

\subsection{Hyperparameter Choices}
\label{appendix:hyperparams}
We follow prior work whenever the recommended settings are available, and otherwise tune on a small held-out subset.
The main hyperparameters for accelerated baselines are:
\begin{itemize}
    \item \textbf{EB-Sampler}: We set the entropy tolerance to $\gamma = 0.01$ for all datasets.
    \item \textbf{Fast-dLLM}: We set the confidence threshold to $\epsilon = 0.95$ for all datasets.
    \item \textbf{KLASS}: We follow the threshold configurations in prior work~\citep{kim2025klass}. KLASS uses two thresholds: a confidence threshold (Conf) and a KL-based threshold (KL). The history length $n$ is set to 2 for all datasets. We report the exact values used for each dataset and model in Table~\ref{tab:klass-exp-config}.
    \item \textbf{DOS}: We report the optimal layer for DOS and DOS$+$EB in Table~\ref{Tab7}.
Entries indicate the transformer layer indices from which the attention matrices are extracted to achieve the best accuracy under each setting.
\end{itemize}

\begin{table}[htbp]
  \centering
  \begin{tabular}{@{}l cc cc@{}}
    \toprule
    Task
      & \multicolumn{2}{c}{LLaDA}
      & \multicolumn{2}{c}{Dream} \\
    \cmidrule(lr){2-3} \cmidrule(lr){4-5}
      & Conf & KL
      & Conf & KL \\
    \midrule
    MATH
      & 0.6   & 0.010
      & 0.9   & 0.005 \\
    GSM8K
      & 0.6   & 0.015
      & 0.9   & 0.001 \\
    HumanEval
      & 0.9   & 0.010
      & 0.8   & 0.001 \\
    MBPP
      & 0.7   & 0.010
      & 0.9   & 0.001 \\
    \bottomrule
  \end{tabular}
  \caption{Threshold configurations for KLASS.}
  \label{tab:klass-exp-config}

\end{table}

\begin{table}[t]
\centering
\small
\resizebox{0.5\textwidth}{!}{
\begin{tabular}{ l| cccc}
\toprule
 &
\textbf{HumanEval}  &
\textbf{MBPP}  &
\textbf{GSM8K} &
\textbf{MATH500}\\

\midrule
\rowcolor{gray!15}
\multicolumn{5}{c}{\textbf{\textit{LLaDA-Instruct-8B}}} \\

DOS (w/o block)
    & 13
    & 9
    & 7
    & 7 \\
DOS (w/ block)
    & 13
    & 9
    & 16
    & 30\\
DOS$+$EB (w/o block)
    & 29
    & 5
    & 4
    & 3 \\
DOS$+$EB (w/ block)
    &  13
    &  1
    &  1
    &  7\\

    \midrule
\rowcolor{gray!15}
\multicolumn{5}{c}{\textbf{\textit{Dream-vo-Instruct-7B}}} \\
DOS (w/o block)
    & 1
    & 22
    & 6
    & 7 \\
DOS(w/ block)
    & 6
    & 6
    & 6
    & 7  \\
DOS$+$EB (w/o block)
    & 6
    & 22
    & 3
    & 3 \\
DOS$+$EB (w/ block)
    & 6
    & 26
    & 3
    & 1 \\
\bottomrule
\end{tabular}}
\caption{
Transformer layers that yield the best performance for DOS under different settings.
For \textbf{DOS}, \emph{w/o block} corresponds to single-block decoding where the block length equals the generation length, while \emph{w/ block} applies block partitioning with a fixed block size of 32.
}
\label{Tab7}
\end{table}

\section{Additional Experiment Results}
\label{B}
\subsection{Experiments on LLaDA}
\label{B.1}
Table~\ref{Tab5} presents the comparative results on HumanEval, MBPP, GSM8K, and MATH500 using the LLaDA-Instruct-8B model with a fixed block size of 32. DOS consistently demonstrates superior generation quality across both Top-1 and parallel sampling settings, proving effective regardless of whether block partitioning is applied. Specifically, in the Top-1 regime, DOS achieves the highest accuracy on all four benchmarks, significantly outperforming standard scoring methods such as Confidence and Entropy. Furthermore, when integrated with the EB-sampler for parallel decoding, DOS maintains this performance advantage, validating the robustness of our scoring strategy in guiding the model toward high-quality outputs.

In terms of computational efficiency, Fast-dLLM exhibits a substantial reduction in the number of function evaluations (NFE), achieving the lowest inference cost across tasks. This efficiency is largely attributed to its threshold-based mechanism. However, this speed comes with a slight trade-off in accuracy compared to our method. Consequently, a promising avenue for future work is to synergize the precise scoring capability of DOS with dynamic threshold-based strategies similar to Fast-dLLM. Such an integration could potentially yield a more optimal generation process, maintaining the high accuracy of DOS while significantly accelerating inference.

\begin{table*}[!htbp]
\centering
\small
\resizebox{\textwidth}{!}{
\begin{tabular}{ l| cc cc cc cc}
\toprule
 &
\multicolumn{2}{c}{\textbf{HumanEval} } &
\multicolumn{2}{c}{\textbf{MBPP} } &
\multicolumn{2}{c}{\textbf{GSM8K}} &
\multicolumn{2}{c}{\textbf{MATH500}} \\
\cmidrule(lr){2-3}
\cmidrule(lr){4-5}
\cmidrule(lr){6-7}
\cmidrule(lr){8-9}
Methods & Acc $\uparrow$ & NFE $\downarrow$
  & Acc $\uparrow$ & NFE $\downarrow$
  & Acc $\uparrow$ & NFE $\downarrow$
  & Acc $\uparrow$ & NFE $\downarrow$  \\

\midrule
\rowcolor{gray!15}
\multicolumn{9}{c}{\textbf{Top-1}} \\

Confidence
    & 40.85 & 256.00
    & 38.20 & 256.00
    & 82.56 & 512.00
    & 38.80 & 512.00 \\
 Entropy
    & 41.46 & 256.00
    & 37.80 & 256.00
    & 82.79 & 512.00
    & 40.00 & 512.00 \\
 Margin
    & 40.85 & 256.00
    & 37.40 & 256.00
    & 82.64 & 512.00
    & 38.80 & 512.00 \\
\textbf{DOS (w/o block) }
    & 42.68 & 256.00
    & 38.40 & 256.00
    & 84.31 & 512.00
    & 41.60 & 512.00 \\
\textbf{DOS (w/ block) }
    & 44.51 & 256.00
    & 38.80 & 256.00
    & 84.61 & 512.00
    & 42.40 & 512.00 \\
    \midrule
\rowcolor{gray!15}
\multicolumn{9}{c}{\textbf{Parallel}} \\

 Fast-dLLM
    & 41.46 & 77.54
    & 38.80 & 54.12
    & 83.40 & 120.56
    & 39.60 & 176.07\\
 KLASS
    & 42.07 & 103.06
    & 38.60 & 74.18
    & 83.02 & 178.26
    & 40.20 & 252.47 \\
 Confidence$+$EB
    & 41.46 & 104.28
    & 38.60 & 73.39
    & 82.71 & 165.33
    & 39.20 & 249.76\\
 Entropy$+$EB
    & 42.68 & 107.88
    & 37.00 & 77.88
    & 82.64 & 176.26
    & 38.60 & 265.76 \\
 Margin$+$EB
    & 40.24 & 102.44
    & 38.80 & 71.31
    & 82.49 & 161.24
    & 37.80 & 243.58\\
\textbf{DOS$+$EB (w/o block) }
    & 45.12 & 115.84
    & 38.80 & 76.09
    & 84.61 & 163.48
    & 41.00 & 244.24 \\
\textbf{DOS$+$EB (w/ block) }
    & 43.90 & 103.05
    & 39.00 & 80.90
    & 84.00 & 170.06
    & 41.40 & 257.55\\
\bottomrule
\end{tabular}}
\caption{
Additional results through top-1 sampling and parallel sampling on HumanEval, MBPP, GSM8K, and MATH500 using \textbf{LLaDA-Instruct-8B} model \citep{nie2025large} with a fixed block size of 32.
\textbf{EB} denotes the EB-sampler \citep{ben2025accelerated}, a parallel decoding method that can be combined with different scoring strategies.
\textbf{Acc} denotes task accuracy, and \textbf{NFE} denotes the number of model forward evaluations.
For HumanEval and MBPP, the generation length is fixed to 256 tokens, while for GSM8K and MATH500, the generation length is 512 tokens.
For \textbf{DOS}, \emph{w/o block} corresponds to single-block decoding, where the block length equals the generation length, while \emph{w/ block} applies block partitioning with block size 32.
}
\label{Tab5}
\end{table*}

\subsection{Experiments on Dream}
\label{B.2}
Table~\ref{Tab6} presents the performance of the Dream-v0-Instruct-7B model across four datasets using both Top-1 and parallel sampling strategies with a fixed block size of 32. Consistent with our previous observations, the proposed DOS method demonstrates superior performance on the majority of tasks with and without block partitioning, particularly excelling in code generation and complex mathematical reasoning. In the Top-1 setting, DOS (w/ block) achieves the highest accuracy on HumanEval and MATH500, significantly outperforming standard baselines like Confidence and Entropy. This advantage extends to the parallel sampling setting, where DOS combined with the EB-sampler attains 45.40\% accuracy on MATH500, surpassing robust competitors such as KLASS and Fast-dLLM.

While DOS remains highly competitive overall, we observe a slight performance gap on the GSM8K benchmark compared to KLASS and uncertainty-based methods, suggesting that our scoring mechanism could be further refined for specific arithmetic reasoning patterns. Furthermore, we note that Fast-dLLM achieves the lowest NFE, demonstrating impressive inference efficiency, while KLASS shows strong adaptability on MBPP; these baselines offer valuable insights into efficiency optimization and task-specific selection, pointing toward promising directions for further enhancing the DOS framework in future work.
\begin{table*}[!htbp]
\centering
\small
\resizebox{\textwidth}{!}{
\begin{tabular}{ l| cc cc cc cc}
\toprule
 &
\multicolumn{2}{c}{\textbf{HumanEval} } &
\multicolumn{2}{c}{\textbf{MBPP} } &
\multicolumn{2}{c}{\textbf{GSM8K}} &
\multicolumn{2}{c}{\textbf{MATH500}} \\
\cmidrule(lr){2-3}
\cmidrule(lr){4-5}
\cmidrule(lr){6-7}
\cmidrule(lr){8-9}
Methods & Acc $\uparrow$ & NFE $\downarrow$
  & Acc $\uparrow$ & NFE $\downarrow$
  & Acc $\uparrow$ & NFE $\downarrow$
  & Acc $\uparrow$ & NFE $\downarrow$  \\

\midrule
\rowcolor{gray!15}
\multicolumn{9}{c}{\textbf{Top-1}} \\

Confidence
    & 58.54 & 256.00
    & 50.02 & 256.00
    & 81.50 & 512.00
    & 40.08 & 512.00 \\
 Entropy
    & 57.32 & 256.00
    & 50.80 & 256.00
    & 83.93 & 512.00
    & 42.20 & 512.00 \\
 Margin
    & 56.71 & 256.00
    & 50.40 & 256.00
    & 81.05 & 512.00
    & 41.80 & 512.00 \\
\textbf{DOS (w/o block) }
    & 59.15 & 256.00
    & 50.60 & 256.00
    & 80.21 & 512.00
    & 45.00 & 512.00 \\
\textbf{DOS (w/ block) }
    & 61.59 & 256.00
    & 50.20 & 256.00
    & 80.29 & 512.00
    & 45.00 & 512.00 \\
    \midrule
\rowcolor{gray!15}
\multicolumn{9}{c}{\textbf{Parallel}} \\

 Fast-dLLM
    & 58.54 & 62.68
    & 50.20 & 50.04
    & 81.50 & 134.17
    & 39.80 & 154.46\\
 KLASS
    & 59.15 & 78.11
    & 52.20 & 72.69
    & 84.08 & 200.85
    & 43.40 & 277.11\\
 Confidence$+$EB
    & 57.93 & 78.86
    & 50.20 & 71.90
    & 81.96 & 182.38
    & 39.80 & 227.54 \\
 Entropy$+$EB
    & 56.10 & 88.29
    & 50.20 & 73.00
    & 83.17 & 192.83
    & 44.20 & 239.13 \\
 Margin$+$EB
    & 56.10 & 80.34
    & 50.20 & 66.90
    & 81.20 & 179.19
    & 39.40 & 226.74 \\
\textbf{DOS$+$EB (w/o block) }
    & 58.54 & 72.47
    & 51.00 & 68.33
    & 79.98 & 182.43
    & 45.40 & 229.36 \\
\textbf{DOS$+$EB (w/ block) }
    & 60.37 & 77.07
    & 50.60 & 71.88
    & 79.83 & 190.20
    & 45.00 & 236.81\\
\bottomrule
\end{tabular}}

\caption{Additional results through top-1 sampling and parallel sampling on HumanEval, MBPP, GSM8K, and MATH500 using \textbf{Dream-v0-Instruct-7B} model \citep{ye2025dream} with a fixed block size of 32.
\textbf{EB} denotes the EB-sampler \citep{ben2025accelerated}, a parallel decoding method that can be combined with different scoring strategies.
\textbf{Acc} denotes task accuracy, and \textbf{NFE} denotes the number of model forward evaluations.
For HumanEval and MBPP, the generation length is fixed to 256 tokens, while for GSM8K and MATH500, the generation length is 512 tokens.
For \textbf{DOS}, \emph{w/o block} corresponds to single-block decoding, where the block length equals the generation length, while \emph{w/ block} applies block partitioning with block size 32.}
\label{Tab6}
\end{table*}

\begin{figure}
    \centering
    \includegraphics[width=\linewidth]{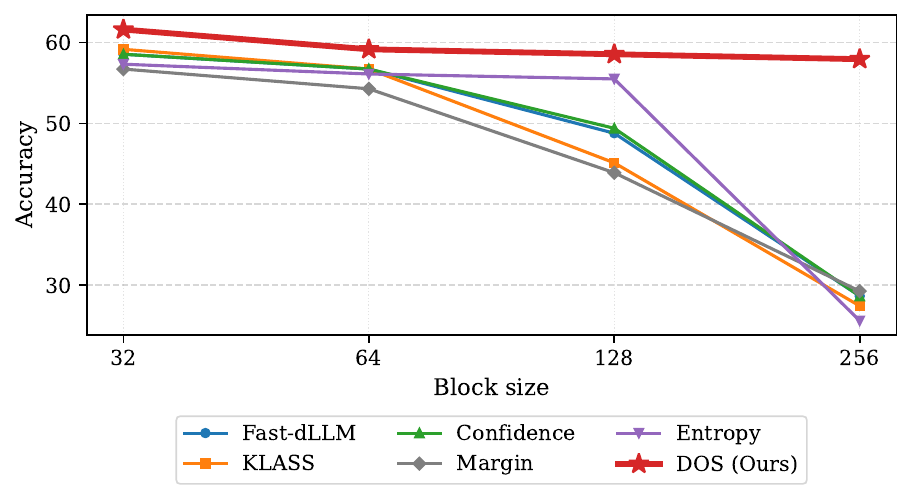}
    \caption{Accuracy on HumanEval using Dream-v0-Instruct-7B with a fixed generation length of 256 under varying block sizes. Existing decoding strategies are sensitive to block size and degrade as the block size increases, whereas DOS (ours) demonstrates strong robustness to block size variation and maintains superior consistency across all settings.}
    \label{fig5}
\end{figure}

Figure~\ref{fig5} shows the performance of different decoding strategies under varying block sizes on the Dream model.
Consistent with the main results, existing uncertainty-based methods exhibit noticeable performance degradation as the block size increases.
In contrast, DOS maintains stable performance across a wide range of block sizes and consistently outperforms competing methods.
These results further confirm the robustness of DOS to block size choices.

\subsection{Performance with a Fixed Attention Layer}
\label{B.3}
To examine whether DOS critically relies on careful attention-layer selection,
we conduct an additional experiment in which the attention matrices are \emph{fixed to the first transformer layer} across all settings.
The results are summarized in Table~\ref{Tab8}.

Across both LLaDA-Instruct-8B and Dream-v0-Instruct-7B, DOS with the first-layer attention consistently achieves performance comparable to the corresponding baselines under all different settings.
This observation holds for both top-1 sampling and parallel decoding with EB-sampler.
With a fixed early-layer attention signal, the proposed dependency-oriented scoring remains sufficient to guide decoding, demonstrating the robustness and practical applicability of DOS.

\begin{table*}[!htbp]
\centering
\small
\resizebox{\textwidth}{!}{
\begin{tabular}{ l| cc cc cc cc}
\toprule
 &
\multicolumn{2}{c}{\textbf{HumanEval} } &
\multicolumn{2}{c}{\textbf{MBPP} } &
\multicolumn{2}{c}{\textbf{GSM8K}} &
\multicolumn{2}{c}{\textbf{MATH500}} \\
\cmidrule(lr){2-3}
\cmidrule(lr){4-5}
\cmidrule(lr){6-7}
\cmidrule(lr){8-9}
Methods & Acc $\uparrow$ & NFE $\downarrow$
  & Acc $\uparrow$ & NFE $\downarrow$
  & Acc $\uparrow$ & NFE $\downarrow$
  & Acc $\uparrow$ & NFE $\downarrow$  \\

\midrule
\rowcolor{gray!15}
\multicolumn{9}{c}{\textbf{\textit{LLaDA-Instruct-8B}}} \\

DOS (w/o block)
    & 41.46 & 256.00
    & 38.20 & 256.00
    & 84.23 & 512.00
    & 41.00 & 512.00 \\
DOS (w/ block)
    & 42.07 & 256.00
    & 37.80 & 256.00
    & 84.23 & 512.00
    & 40.40 & 512.00 \\
DOS$+$EB (w/o block)
    & 43.90 & 104.93
    & 38.60 & 76.34
    & 83.55 & 164.26
    & 39.60 & 258.37 \\
DOS$+$EB (w/ block)
    & 40.85 & 107.11
    & 39.00 & 80.90
    & 84.00 & 170.06
    & 40.60 & 260.90 \\
    \midrule
\rowcolor{gray!15}
\multicolumn{9}{c}{\textbf{\textit{Dream-vo-Instruct-7B}}} \\
DOS (w/o block)
    & 59.15 & 256.00
    & 49.80 & 256.00
    & 79.53 & 512.00
    & 44.20 & 512.00 \\
DOS (w/ block)
    & 59.15 & 256.00
    & 49.80 & 256.00
    & 79.53 & 512.00
    & 44.20 & 512.00\\
DOS$+$EB (w/o block)
    & 57.32 & 75.49
    & 50.00 & 69.05
    & 79.83 & 181.63
    & 44.80 & 230.09 \\
DOS$+$EB (w/ block)
    & 57.32 & 79.44
    & 50.00 & 73.34
    & 79.83 & 190.81
    & 45.00 & 236.81\\
\bottomrule
\end{tabular}}
\caption{
Results of DOS and DOS$+$EB with attention matrices extracted from the first transformer layer.
Experiments are conducted on LLaDA-Instruct-8B and Dream-v0-Instruct-7B under both single-block and block-based decoding settings.
For HumanEval and MBPP, the generation length is fixed to 256 tokens, while for GSM8K and MATH500, the generation length is 512 tokens.
\textbf{DOS} \emph{w/o block} denotes single-block decoding where the block length equals the generation length, while \textbf{DOS} \emph{w/ block} applies block partitioning with a fixed block size of 32.
\textbf{Acc} denotes task accuracy, and \textbf{NFE} denotes the number of model forward evaluations, reflecting decoding efficiency.
}
\label{Tab8}
\end{table*}
\clearpage
\newpage

\begin{table*}[!htbp]
\centering
\small
\resizebox{\textwidth}{!}{
\begin{tabular}{ l| cccc | cccc | cccc | cccc}
\toprule
 &
\multicolumn{4}{c}{\textbf{HumanEval} } &
\multicolumn{4}{c}{\textbf{MBPP} } &
\multicolumn{4}{c}{\textbf{GSM8K}} &
\multicolumn{4}{c}{\textbf{MATH500}} \\
\cmidrule(lr){2-5}
\cmidrule(lr){6-9}
\cmidrule(lr){10-13}
\cmidrule(lr){14-17}
Methods & Acc $\uparrow$ & NFE $\downarrow$ & TPS $\uparrow$ & Time $\downarrow$
  & Acc $\uparrow$ & NFE $\downarrow$ & TPS $\uparrow$ & Time $\downarrow$
  & Acc $\uparrow$ & NFE $\downarrow$ & TPS $\uparrow$ & Time $\downarrow$
  & Acc $\uparrow$ & NFE $\downarrow$ & TPS $\uparrow$ & Time $\downarrow$  \\

\midrule
\rowcolor{gray!15}
\multicolumn{17}{c}{\textbf{Top-1}} \\

Confidence
    & 40.85 & 256.00 & 4.22 & 32.40
    & 38.20 & 256.00 & 5.07 & 20.21
    & 82.56 & 512.00 & 4.96 & 66.22
    & 38.80 & 512.00 & 6.46 & 67.59\\
 Entropy
    & 41.46 & 256.00 & 4.28 & 33.07
    & 37.80 & 256.00 & 5.04 & 20.62
    & 82.79 & 512.00 & 4.78 & 68.37
    & 40.00 & 512.00 & 6.28 & 69.57 \\
 Margin
    & 40.85 & 256.00 & 4.07 & 33.10
    & 37.40 & 256.00 & 4.84 & 20.56
    & 82.64 & 512.00 & 4.78 & 68.01
    & 38.80 & 512.00 & 6.34 & 69.35 \\
\textbf{DOS (w/o block) }
    & 42.68 & 256.00 & 4.71 & 32.18
    & 38.40 & 256.00 & 5.76 & 19.84
    & 84.31 & 512.00 & 5.00 & 65.47
    & 41.60 & 512.00 & 6.56 & 66.85\\
\textbf{DOS (w/ block) }
    & 44.51 & 256.00 & 4.78 & 32.26
    & 38.80 & 256.00 & 5.78 & 19.87
    & 84.61 & 512.00 & 4.88 & 65.39
    & 42.40 & 512.00 & 6.60 & 66.72\\
    \midrule
\rowcolor{gray!15}
\multicolumn{17}{c}{\textbf{Parallel}} \\

 Fast-dLLM
    & 41.46 & 77.54 & 13.68 & 10.00
    & 38.80 & 54.12 & 24.19 & 4.32
    & 83.40 & 120.56 & 21.01 & 15.85
    & 39.60 & 176.07 & 19.59 & 23.62 \\
 KLASS
    & 42.07 & 103.06 & 8.79 & 15.51
    & 38.60 & 74.18 & 15.66 & 6.53
    & 83.02 & 178.26 & 12.58 & 26.23
    & 40.20 & 252.47 & 11.91 & 37.94 \\
 Saber
    & 43.29 & 67.15 & 17.32 & 8.49
    & 36.60 & 53.71 & 26.20 & 4.19
    & 82.64 & 123.35 & 20.81 & 15.79
    & 40.00 & 176.03 & 19.80 & 23.01\\

 WINO
    & 40.85 & 124.73 & 7.84 & 18.03
    & 38.60 & 94.01 & 12.50 & 8.22
    & 77.33 & 312.18 & 7.29 & 43.91
    & 26.80 & 363.71 & 8.52 & 51.50 \\
 Confidence$+$EB
    & 41.46 & 104.28 & 10.28 & 13.42
    & 38.60 & 73.39 & 18.24 & 5.71
    & 82.71 & 165.33 & 15.66 & 21.24
    & 39.20 & 249.76 & 13.90 & 32.80\\
 Entropy$+$EB
    & 42.68 & 107.88 & 10.33 & 14.12
    & 37.00 & 77.88 & 17.40 & 6.12
    & 82.64 & 176.26 & 14.65 & 22.73
    & 38.60 & 265.76 & 13.01 & 34.94 \\
 Margin$+$EB
    & 40.24 & 102.44 & 10.26 & 13.29
    & 38.80 & 71.31 & 18.16 & 5.61
    & 82.49 & 161.24 & 15.87 & 20.81
    & 37.80 & 243.58 & 14.12 & 32.12 \\

\textbf{DOS$+$EB (w/o block) }
    & 45.12 & 115.84 & 11.38 & 15.02
    & 38.80 & 76.09  & 18.20 & 6.14
    & 84.61 & 163.48 & 15.99 & 21.17
    & 41.00 & 244.24 & 14.32 & 32.35\\
\textbf{DOS$+$EB (w/ block) }
    & 43.90 & 103.05 & 12.25 & 13.39
    & 39.00 & 80.90 & 17.16 & 6.35
    & 84.00 & 170.06 & 15.36 & 21.74
    & 41.40 & 257.55 & 13.63 & 33.49\\
\bottomrule
\end{tabular}}
\caption{
Additional results through top-1 sampling and parallel sampling on HumanEval, MBPP, GSM8K, and MATH500 using \textbf{LLaDA-Instruct-8B} model \citep{nie2025large} with a fixed block size of 32.
\textbf{EB} denotes the EB-sampler \citep{ben2025accelerated}, a parallel decoding method that can be combined with different scoring strategies.
\textbf{Acc} denotes task accuracy, and \textbf{NFE} denotes the number of model forward evaluations.
For HumanEval and MBPP, the generation length is fixed to 256 tokens, while for GSM8K and MATH500, the generation length is 512 tokens.
For \textbf{DOS}, \emph{w/o block} corresponds to single-block decoding, where the block length equals the generation length, while \emph{w/ block} applies block partitioning with block size 32.
}

\end{table*}
\end{document}